# Fast Convolution Based on Winograd Minimum Filtering: Introduction and Development

Gan Tong and Libo Huang

School of Computer, National University of
Defense Technology, Changsha, China

## Abstract

*Convolutional Neural Network (CNN) has been widely used in various fields and played an important role. Convolution operators are the fundamental component of convolutional neural networks, and it is also the most time-consuming part of network training and inference. In recent years, researchers have proposed several fast convolution algorithms including FFT and Winograd. Among them, Winograd convolution significantly reduces the multiplication operations in convolution, and it also takes up less memory space than FFT convolution. Therefore, Winograd convolution has quickly become the first choice for fast convolution implementation within a few years. At present, there is no systematic summary of the convolution algorithm. This article aims to fill this gap and provide detailed references for follow-up researchers. This article summarizes the development of Winograd convolution from the three aspects of algorithm expansion, algorithm optimization, implementation, and application, and finally makes a simple outlook on the possible future directions.*

## 1. Introduction

Convolutional Neural Networks (CNN) are widely used in tasks such as computer vision and natural language processing. CNN with deep learning has reached or even surpassed the level of human experts in some fields by constructing deeper and more complex networks. At the same time, deeper CNN also brings more parameters and greater computing power requirements. Therefore, more and more research attempts to accelerate the training and inference of CNN and the use of fast convolution operators is an important method among them.

Fast convolution operator uses fast convolution algorithm to achieve convolution, including FFT convolution [1] and Winograd convolution [2]. This type of convolution converts the matrix multiplication in the original convolution into the corresponding element-wise multiplication (Hadamard product) by linearly transforming the input feature map and convolution kernel of the convolution operator to the corresponding domain. The result of the corresponding element-wise multiplication can be restored to the original feature mapping domain after the corresponding inverse linear transformation. In this "transform-calculation-inverse transformation" process, the number of multiplication operations is considerably reduced compared to direct convolution, and the cost is an increase in the number of addition operations.

On most modern processors, the execution efficiency of addition is much higher than that of multiplication. Therefore, we can replace the convolution implementation in the model with a fast convolution operator and use the reduced multiplication operations to improve the execution efficiency of the model. The same is to reduce the multiplication operation. The linear transformation in the Winograd convolution is to map the real number to the real number domain, and the FFT convolution is to map to the complex number domain. Therefore, the memory usage during the Winograd convolution operation only needs half of the FFT convolution, making the Winograd convolution the most popular fast convolution operator.

However, there are many challenges in directly applying Winograd convolution. First, the earliest proposed Winograd convolution has a limited scope of application. It can only be applied to two-dimensional convolutions with unit stride size and small convolution kernels. When applied to large convolution kernels, there will be numerical instability [3]. Secondly, due to the complexity of linear transformation and inverse linear transformation, the optimization of fast convolution operators on a specific platform is difficult to achieve, such as the use of parallelism and data locality [4]. In addition, Winograd convolution and network compression technology represented by pruning and quantization are difficult to directly combine, so it is not easy to be deployed on platforms with insufficient computing power and energy consumption restrictions [5].

In response to these problems, researchers have done a lot of work, but so far there is no published article that systematically summarizes related work. In order to make it easier for the follow-up researchers to understand and master the previous work, this article summarizes the development of Winograd from the three aspects of algorithm generalization, algorithm optimization, implementation and application, and looks forward to the possible future directions. The structure of this paper is as follows: Section 2 introduces the introduction and algorithm development of Winograd convolution; Section 3 introduces the optimization of Winograd convolution algorithm in three aspects; Section 4 introduces the realization and practical application of Winograd convolution on several types of platforms; Chapter Five summarizes this article and looks forward to possible future research directions.

## 2. THE INTRODUCTION AND DEVELOPMENT OF WINOGRAD CONVOLUTION

### 2.1. Winograd Minimum Filtering Algorithm

Winograd proposed the minimum filtering algorithm of finite impulse response (FIR) filtering in 1980 [6]. The minimum filtering algorithm gives $m$ output generated by the FIR filter of $r$ taps, namely $F(m, r)$, the minimum number of multiplications needed $\mu(F(m,, r))$ is $m + r - 1$. Taking $F(2,3)$ as an example, the input $d = [d_0, d_1, d_2, d_3]$ is a vector of size 4, filter $g = [g_0, g_1, g_3]^T$, then:

$$F(2,3) = \begin{bmatrix} d_0 & d_1 & d_2 \\ d_1 & d_2 & d_3 \end{bmatrix} \begin{bmatrix} g_0 \\ g_1 \\ g_2 \end{bmatrix} = \begin{bmatrix} m_1 + m_2 + m_3 \\ m_2 - m_3 - m_4 \end{bmatrix}$$

where:

$$m_1 = (d_0 - d_2)g_0 \quad m_2 = (d_1 + d_2)\frac{g_0 + g_1 + g_2}{2}$$
$$m_4 = (d_1 - d_3)g_2 \quad m_3 = (d_2 - d_1)\frac{g_0 - g_1 + g_2}{2}$$

When calculating $[m_1, m_2, m_3, m_4]$, the number of multiplications involved in the algorithm is $\mu(F(2,3)) = 2 + 3 - 1 = 4$, the number of addition operations that need to be performed on $d$ is 4, and the number of addition operations that need to be performed on $g$ is 3 (the value of $g_0 + g_2$ can be calculated only once); using $[m_1, m_2, m_3, m_4]$ to get the result of $F(2,3)$ requires 4 additions. The number of multiplications required by the algorithm has been reduced from 6 to 4.

## 2.2. The Introduction of Winograd convolution

Winograd minimum filtering algorithm can be expressed in the form of a matrix:

$$Y = A^T[(Gg) \odot (B^T d)]$$

where $g$ is the filter vector, $d$ is the input data vector, $Y$ is the output data vector, $G$ is the filter transformation matrix, $B^T$ is the data transformation matrix, $\odot$ is the corresponding bit multiplication of the matrix (Hadamard product), $A^T$ represents the output transformation matrix. For $F(2,3)$, the matrices are:

$$B = \begin{bmatrix} 1 & 0 & -1 & 0 \\ 0 & 1 & 1 & 0 \\ 0 & -1 & 1 & 0 \\ 0 & 1 & 0 & -1 \end{bmatrix}, \quad G = \begin{bmatrix} 1 & 0 & 0 \\ \frac{1}{2} & \frac{1}{2} & \frac{1}{2} \\ \frac{1}{2} & -\frac{1}{2} & \frac{1}{2} \\ 0 & 0 & 1 \end{bmatrix},$$

$$A^T = \begin{bmatrix} 1 & 1 & 1 & 0 \\ 0 & 1 & -1 & -1 \end{bmatrix}, g = [g_0 \quad g_1 \quad g_2]^T, d = [d_0 \quad d_1 \quad d_2 \quad d_3]^T$$

By nesting the one-dimensional minimum filtering algorithm $F(m,r)$, we can get the two-dimensional minimum filtering algorithm $F(m \times m, r \times r)$:

$$Y = A^T[(GgG^T) \odot (B^T d B)]A$$

Now the size of the filter $g$ is $r \times r$, the size of output $Y$ is $m \times m$, and the size of input $d$ is $(m + r - 1) \times (m + r - 1)$. The number of multiplications required by the two-dimensional minimum filtering algorithm is $(m + r - 1)^2$, while the number of multiplications required by the original convolution algorithm is $m \times m \times r \times r$. For $F(2 \times 2, 3 \times 3)$, the number of multiplications is reduced from 36 to 16, which is a reduction of 2.25 times. Even if the additional addition operations are included, there are great benefits. We can naturally split Winograd convolution into four separate stages (as shown in Figure 1):

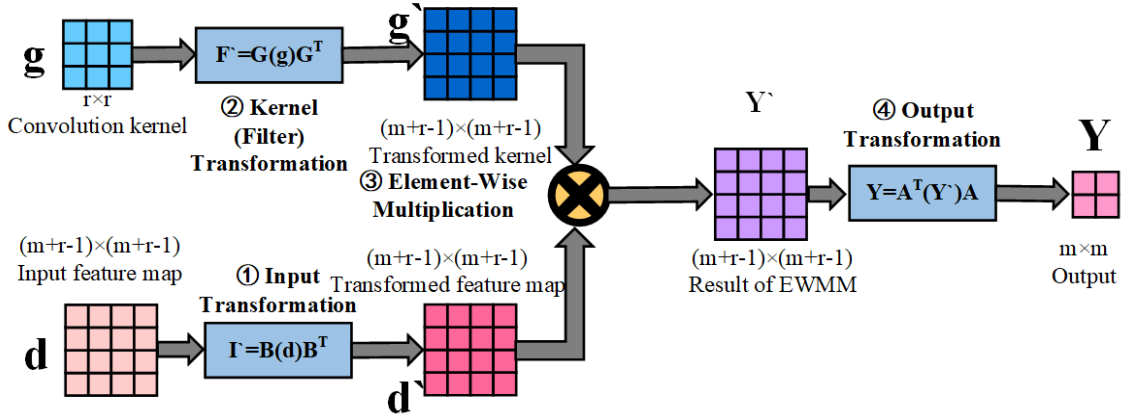

Figure 1. Four stages of Winograd convolution

- **Input Transformation** (**ITrans**): $d' = B^T dB$, transform the input tensor to Winograd domain, the size of $d'$ is $(m + r - 1) \times (m + r - 1)$;
- **Kernel Transformation** (**KTrans**): $g' = GgG^T$, transform the convolution kernel to Winograd domain, the size of $g'$ is $(m + r - 1) \times (m + r - 1)$;
- **Element-Wise Matrix Multiplication** (**EWMM**): $Y' = g' \odot d'$, which is the calculation stage of Winograd convolution, the size of $Y'$ is $(m + r - 1) \times (m + r - 1)$;
- **Output Transformation** (**OTrans**): $Y = A^T Y' A$, inversely transform the result of EWMM from Winograd domain to feature map tensor domain, the size of Y is $m \times m$.

[2] was the first that applied the Winograd minimum filtering algorithm to CNN, and the performance of the convolution operator is improved by reducing the number of multiplications. For a two-dimensional convolution operator, the output needs to be divided into $m \times m$ tiles. The input corresponding to the convolution is the input slices of $(m + r - 1) \times (m + r - 1)$ that overlap each other. There is an overlap of $r - 1$ between the input slice and the adjacent slice. According to the analysis of [7], a large slice size and a small convolution kernel size can make the overlap area of repeated calculations less, but it will also bring greater numerical errors. Experiments show that the performance of $F(2 \times 2, 3 \times 3)$ on multiple convolutions exceeds cuDNN, and the memory size used is much lower than FFT convolution. As a result, the Winograd convolution was introduced into CNN.

## 2.3. Generalization and Extension of Winograd Convolution

### 2.3.1. Generalization

The introduction of Winograd convolution [2] is a milestone, but the convolution only supports two-dimensional convolution operators with $r = 3$ and $r = 2$, and the tilesize does not exceed 6. But this is far from satisfying the various types of convolution operators in modern CNNs, so there are many subsequent studies that generalize Winograd convolution to various types.

[8], [9], [10] realized Winograd convolution to support larger arbitrary convolution kernel size and slice size, [11], [12], [13], [14] proposed a decomposition method to decompose the large convolution kernel size and tile size into several kind of small Winograd convolution. Three-dimensional convolution is used to process time and space feature information and is the main component of 3D-CNN. [14], [15], [16], [17], [18]nested one-dimensional Winograd convolution

and two-dimensional Winograd convolution to obtained three-dimensional Winograd convolution. [8] generalized the Winograd convolution to N-dimensions. Down-sampling often uses a stride convolution operator to reduce the size of the feature map through non-unit step convolution. [19] extended the algorithm to three dimensions while achieving a step size of 2. [10], [11], [12], [13], [14] used matrix decomposition to extend the algorithm to any stride size. [20] combined the decomposition algorithm and nesting to better solve the problem of arbitrary stridesize. Dilated convolution and transposed convolution are often used in image segmentation, super-resolution and other fields. [21] proposed a dilated Winograd convolution to support the dilation of 2 and 4. [22] converted the transposed convolution into multiple basic convolutions through predefined decomposition and interleaving operations and implemented the support of the Winograd convolution for the transposed convolution.

### 2.3.2. Extension

In addition to generalizing to various convolutions, there are many attempts to extend the linear transformation of Winograd convolution itself. The Winograd algorithm family linearly transforms the input tiles and convolution kernel into the Winograd domain, performs the Hadamard product, and then inversely transforms back to the feature map domain. For the specified convolution kernel and tile size, the linear transformation matrices $A$, $G$, and $B$ are known before calculation. Convolution can be expressed as polynomial multiplication. Map the elements of the convolution kernel $g(x)$ and the input vector $d(x)$ to the coefficients of the polynomials $g(x)$ and $d(x)$ respectively, then the elements of the output vector $y$ (convolution of $g$ and $d$) are equal to the coefficient of the polynomial $y(x) = g(x)d(x)$. The Winograd convolution algorithm family is based on the Chinese Remainder Theorem (CRT) on polynomials. The convolution output can be obtained by taking the remainder of the polynomial in the irreducible and coprime polynomial congruence system. Solving the congruence equations is to obtain the specific solution of the linear transformation matrix according to the coefficients of the polynomial [6].

[23] extended the convolution polynomial used in the Winograd convolution algorithm to a higher-order polynomial. Experiments have shown that second-order polynomials will significantly reduce the error, but will also increase the number of multiplications, so there is a trade-off between the number of multiplications and the accuracy of floating-point numbers. [24] extended polynomial multiplication to the complex number domain and used the symmetry of conjugate complex multiplication to further reduce the number of multiplications. [25] proposed to introduce Winograd convolution into the remainder system (RNS) to realize the low-precision quantization operation of Winograd convolution and support larger input slice size. [26] Innovatively introduced the Fermat Number Transformation (FNT). On the one hand, using this transformation can ensure that the intermediate calculation results are all unsigned numbers, and on the other hand, all calculations are simplified to shift and addition operations.

In addition, [27] combined Winograd convolution with Strassen's algorithm. Strassen algorithm [28] is an algorithm to reduce the number of matrix operations. [2] pointed out that the operation reduction of Strassen algorithm is much smaller than Winograd algorithm, but [27] replaced the convolution used in Strassen algorithm with Winograd convolution and combined the reduction of operations brought by the two to achieve further optimization. [29] applied Winograd convolution to the additive neural network, replacing multiplication with addition, maintaining considerable performance and reducing power consumption.

## 3. OPTIMIZATION OF WINOGRAD CONVOLUTION

### 3.1. Pruning

Pruning is an effective technique commonly used in CNN optimization. Pruning is mainly used to prune the weights of the convolution operator in CNN, and the weights that have little effect on the output will be set to zero. The convolution kernel after pruning becomes a sparse tensor, which brings two advantages. One is that storing sparse convolution kernel tensor weights in a specific compression format can reduce memory usage, and the other is that many elements in the sparse tensor are 0, so the amount of calculation for convolution can be reduced. For the convolution of the convolutional layer and the fully connected layer, the parameters can be reduced by more than 90%. However, it is difficult to directly apply pruning on Winograd convolution, because the sparse convolution kernel will become a dense matrix after transforming to the Winograd domain, which violates the original intention of pruning.

[5], [30], [31], [32] propose to apply pruning on Winograd convolution and FFT convolution. They applied a linear rectification unit (ReLU) after the input transformation to obtain a sparse Winograd domain tensor. At the same time, the transformed convolution kernel is pruned to obtain a sparse Winograd domain convolution kernel. At this point, the two tensors in the calculation phase have become sparse tensors. [33] applied pooling after the input transformation, the principle is the same as the application of ReLU. [34], [35] designed a new memory data layout for sparse Winograd convolution. [36] proposed to learn the pruning coefficient of Winograd convolution locally and reached a sparse rate of more than 90%. [37] pointed out that the use of ReLU method changed the network layout, they proposed to apply spatial structure pruning on the transformed feature map tensor, and then transfer its sparsity to the convolution kernel of the Winograd domain. [10], [38] proposed that the results of the pruning of the above methods are irregular, which is not conducive to the performance of hardware, so the position-sensitive sub-row balance coefficient pruning mode and sparse row balance compression are respectively proposed. [39] proposed a new coding format to solve the coding overhead caused by the sparsity of the active area. [40] introduced the Zero-Skip hardware mechanism, skipped the calculation of 0 weight, and provided hardware support for the sparse matrix operation after pruning.

### 3.2. Low Precision and Quantization

It is another common method in CNN to sacrifice precision and reduce memory footprint and computational efficiency. Changing the parameters of the CNN model from a 32-bit floating-point number to a 16-bit floating-point number or quantizing it to an 8-bit fixed-point number or even lower precision, can compress the model and improve computing efficiency without losing the accuracy of the model. When Winograd convolution was first introduced, single-precision and half-precision floating-point numbers were tested at the same time, but experiments have shown that using half-precision floating-point numbers will lead to larger absolute errors [2]. Winograd convolution can also be combined with quantization. [24] proposed a uniform affine quantization to generate a quantized convolution kernel and expressed in 8-bit unsigned integer and dynamic range. [41] proposed dynamic layered application of different convolution implementation and quantization on CNN to reduce computational complexity, including the quantization of Winograd convolution. [42] proposed to apply Winograd convolution to an 8-bit network and use learning to solve the problem of accuracy loss. [25] The introduction of RNS transformation also enables quantization to operate at low precision. [43] proposed to model the accuracy loss and use different quantization levels for feature maps and convolution kernels. [44] replaced the canonical basis polynomials in the Winograd transform with Legendre basis polynomials, and proposed quantization based on basis transformation techniques. [45] embed linear quantization directly into the Winograd domain to achieve low-precision quantization. [46] further explored the use of the Winograd algorithm to optimize the convolution kernel with 4-6-bit precision. [47] applied quantization on feature map slices, and applied particle swarm optimization technology to find the threshold of quantization. In addition, [10], [48], [49], [50] also applied 8-bit quantization technique on Winograd convolution.

### 3.3. Numerical Stability

Winograd convolution has only been applied to the $3 \times 3$ convolution kernel and small input tiles for a long time, because of the inherent numerical instability in the Winograd convolution calculation. When applied to larger convolution kernels or input tiles, the polynomial coefficients of the Winograd transform increase exponentially. This imbalance will be reflected in the elements of the transformation matrix, resulting in large relative errors. [7] studied that the source of this numerical instability is the large-scale Vandermonde matrix in the transformation [3] and proposed carefully selecting the corresponding polynomials that exhibit the smallest exponential growth. They also proposed scaling the transformation matrix to alleviate numerical instability. [23] used higher-order polynomials to reduce the error of Winograd convolution, but the cost was an increase in the number of multiplications. [42] handed over the processing of numerical errors to training to learn better convolution kernel weights and quantization in Winograd convolution. [51] proved mathematically that large convolution kernels can be solved by overlap and addition. [20], [52] solved large-size convolution kernel and non-unit step convolution into small convolution kernels to solve the numerical accuracy problem. [53] selected the appropriate output tile size based on symbolic calculation and meta-programming automation to balance numerical stability and efficiency. [54] proved that the floating-point calculation order in linear transformation affects accuracy, rearranged the calculation order in linear transformation based on Huffman coding, and proposed a mixed-precision algorithm.

## 4. IMPLEMENTATIONS AND APPLICATIONS OF WINOGRAD CONVOLUTION

### 4.1. Implementation

The high performance brought by Winograd convolution allows researchers to quickly deploy it to various platforms, in addition to CPUs, GPUs, etc., but also FPGA platforms, mobile terminals, and edge computing devices that have strict requirements for efficiency and power consumption.

#### 4.1.1. CPU

[5], [36] implemented pruning and retraining of Winograd convolution on the CPU. [55], [56] compared the performance of FFT and Winograd on CPU, and their performance characteristics were analysed. [15] implemented three-dimensional Winograd convolution on a multi-core CPU by using the specific large memory of the CPU platform. [57] use JIT optimization technology to accelerate the realization of the direct convolution kernel Winograd convolution on the small convolution kernel on the x86 CPU architecture. [58] relied on the automatic vectorization of the compiler, the calculation stage was converted to batched GEMM to achieve performance improvement. [8] proposed a custom data layout on the CPU, using vectorized instructions to achieve efficient memory access. [59] used the L3-Cache of the CPU to reuse the convolution kernel, but it cannot support convolution with an excessively large number of channels. [60] utilized the similarity in Winograd convolution to achieve deep data reuse on the CPU. In

addition, [61], [62] implemented Winograd convolution and efficient inference libraries on the mobile ARM platform CPU.

### 4.1.2. GPU

[63] performed similar performance comparisons with [55], [56] on the GPU. [16], [17] implemented the three-dimensional Winograd convolution on the GPU, but the calculation stage in [16] directly called the matrix multiplication implementation of the cuBLAS library, and [17] manually wrote a specific implementation. The data parallelism and intra-slice parallelism of Winograd convolution are used in [64] to achieve multi-dimensional parallel training on large-scale GPU clusters. [65] used MegaKernel technique to fuse the four stages of Winograd convolution and used a well-designed task mapping algorithm to achieve a significant performance improvement on the GPU. [66] used SASS assembler to optimize Winograd convolution, merge global memory access and make shared memory access conflict-free, used cache to design pipeline and to improve calculation intensity, and used conventional registers to fill the shortcomings of insufficient predicate registers. [32] realized the pruning technology on the GPU and proposed a dynamic batch size algorithm to improve the training speed. [53] also implemented Winograd convolution for mobile GPUs.

### 4.1.3. FPGA

[67] described in detail the minimum requirements for Winograd convolution hardware implementation and implemented the basic modules of Winograd convolution on FPGA. [68] cached all intermediate feature mappings in the stream buffer to achieve a high energy consumption ratio FPGA implementation. [69] designed a line buffer structure to cache feature maps and reuse data from different slices, and designed an efficient parallel execution unit for Winograd convolution, and dealt with the sparse case in [34]. [14], [70], [71] unified the two-dimensional and three-dimensional Winograd, and built a unified template on the FPGA. [72] implemented hybrid convolution on FPGA and analysed the occasions suitable for FFT and Winograd convolution. [35], [73], [74], [75] unified the realization of the Winograd convolution kernel matrix multiplication and maximize the reusability of the module. [76], [77] conducted a comprehensive design space exploration on the realization of Winograd convolution on FPGA. [11] proposed a decomposition method for the convolution of various parameters, which simplifies the hardware implementation. [78] designed an open-source back-end framework on the CPU-FPGA heterogeneous platform, but only supports Winograd convolution with unit stride size, while [43] implements fine-grained scheduling and supports more general convolution. In addition, [79], [80], [81], [82], [83] and [84] also implemented Winograd convolution on FPGA and explored the design space, and [85] fully evaluated the complete CNN implementation on FPGA.

### 4.1.4. Other

[40], [86] utilized the SIMT architecture of GPGPU to process CNN using Winograd convolution in parallel, and [40] also added support for Zero-Skip. [87], [88] used high-efficiency Winograd convolution on IoT devices to achieve high performance. [41], [89], [90] used random calculation and approximate calculation to complete the implementation. [91] is implemented on ReRAM, which improves data reuse based on tiles, and [48] implemented 8-bit quantized convolution based on DRAM architecture. [18] realized the three-dimensional Winograd convolution on the vector DSP. By expanding a $F(2 \times 2, 3 \times 3)$ convolution instruction and adding a new calculation module, [92] implemented Winograd convolution on an open source RISC-V framework.

Many frameworks integrate Winograd convolution to improve model execution efficiency. Cltorch [4] is a hardware-independent back-end platform based on OpenCL. [93] implemented a tool for mapping the Caffe model to FPGA and chose whether to apply Winograd convolution based on dynamic programming. [94] implemented a software stack supporting Winograd convolution, generating high-efficiency load for Intel hardware including CPU, integrated display, and neural computing stick. In addition, Winograd convolution has been integrated into popular deep learning frameworks and neural network libraries.

### 4.2. Applications

The fast convolution represented by Winograd aims to accelerate the convolution to improve the execution efficiency of the CNN model. It can be used in scenarios that require real-time performance. [72] implemented a face recognition system using hybrid convolution. [95] Implemented an accelerator for action recognition based on three-dimensional Winograd convolution. [22,96] introduced Winograd convolution into real-time super-resolution, but there is a difference in upsampling. [22] uses the Winograd implementation of transposed convolution, while [96] uses shuffle layer instead of transposed convolution. [49] implemented a speech recognition accelerator on wearable devices and applied an 8-bit integer Winograd convolutional network.

## 5. CONCLUSION

Winograd convolution is currently the most widely used fast convolution operator. Since the introduction of CNN, its scope of use has gradually covered all types of convolutions in modern CNN with the in-depth research of researchers, and the combination with pruning, quantization and other technologies has also matured. Winograd convolution has been integrated in various platform deep learning frameworks and neural network libraries, which can generate efficient workloads for various hardware platforms. On the FPGA platform, it is possible to customize the implementation of software and hardware coordination for Winograd convolution, but how to make good use of computing power and memory levels on general computing platforms such as CPU and GPU remains to be further studied. For example, the four-stage integration of Winograd, the optimization of data flow, the trade-off between computational intensity and memory access efficiency, may all be a breakthrough in optimization on a general-purpose computing platform.

188	Computer Science & Information Technology (CS & IT)[44] B. Barabasz, "Quantaized Winograd/Toom-Cook Convolution for DNNs: Beyond Canonical Polynomials Base," ArXiv200411077 Cs Math Stat, Apr. 2020, Accessed: Sep. 19, 2021. [Online]. Available: http://arxiv.org/abs/2004.11077

[45] G. Li, L. Liu, X. Wang, X. Ma, and X. Feng, "Lance: efficient low-precision quantized winograd convolution for neural networks based on graphics processing units," in ICASSP 2020 - 2020 IEEE International Conference on Acoustics, Speech and Signal Processing (ICASSP), Barcelona, Spain, May 2020, pp. 3842–3846. doi: 10.1109/ICASSP40776.2020.9054562.

[46] Q. Han et al., "Extremely Low-bit Convolution Optimization for Quantized Neural Network on Modern Computer Architectures," in 49th International Conference on Parallel Processing - ICPP, Edmonton AB Canada, Aug. 2020, pp. 1–12. doi: 10.1145/3404397.3404407.

[47] D. Sabir, M. A. Hanif, A. Hassan, S. Rehman, and M. Shafique, "TiQSA: Workload Minimization in Convolutional Neural Networks Using Tile Quantization and Symmetry Approximation," IEEE Access, vol. 9, pp. 53647–53668, 2021, doi: 10.1109/ACCESS.2021.3069906.

[48] M. M. Ghaffar, C. Sudarshan, C. Weis, M. Jung, and N. Wehn, "A Low Power In-DRAM Architecture for Quantized CNNs using Fast Winograd Convolutions," in The International Symposium on Memory Systems, Washington DC USA, Sep. 2020, pp. 158–168. doi: 10.1145/3422575.3422790.

[49] Y. Yao et al., "INT8 Winograd Acceleration for Conv1D Equipped ASR Models Deployed on Mobile Devices," ArXiv201014841 Cs Eess, Oct. 2020, Accessed: Sep. 19, 2021. [Online]. Available: http://arxiv.org/abs/2010.14841

[50] Y. Cao, C. Song, and Y. Tang, "Efficient LUT-based FPGA Accelerator Design for Universal Quantized CNN Inference," in 2021 2nd Asia Service Sciences and Software Engineering Conference, Macau Macao, Feb. 2021, pp. 108–115. doi: 10.1145/3456126.3456140.

[51] C. Ju and E. Solomonik, "Derivation and Analysis of Fast Bilinear Algorithms for Convolution," SIAM Rev., vol. 62, no. 4, pp. 743–777, Jan. 2020, doi: 10.1137/19M1301059.

[52] D. Huang et al., "DWM: A Decomposable Winograd Method for Convolution Acceleration," Proc. AAAI Conf. Artif. Intell., vol. 34, no. 04, pp. 4174–4181, Apr. 2020, doi: 10.1609/aaai.v34i04.5838.

[53] A. Mazaheri, T. Beringer, M. Moskewicz, F. Wolf, and A. Jannesari, "Accelerating winograd convolutions using symbolic computation and meta-programming," in Proceedings of the Fifteenth European Conference on Computer Systems, Heraklion Greece, Apr. 2020, pp. 1–14. doi: 10.1145/3342195.3387549.

[54] B. Barabasz, A. Anderson, K. M. Soodhalter, and D. Gregg, "Error Analysis and Improving the Accuracy of Winograd Convolution for Deep Neural Networks," ACM Trans. Math. Softw., vol. 46, no. 4, pp. 1–33, Nov. 2020, doi: 10.1145/3412380.

[55] A. Zlateski, Z. Jia, K. Li, and F. Durand, "FFT Convolutions are Faster than Winograd on Modern CPUs, Here is Why," ArXiv180907851 Cs, Sep. 2018, Accessed: Sep. 19, 2021. [Online]. Available: http://arxiv.org/abs/1809.07851

[56] A. Zlateski, Z. Jia, K. Li, and F. Durand, "The anatomy of efficient FFT and winograd convolutions on modern CPUs," in Proceedings of the ACM International Conference on Supercomputing, Phoenix Arizona, Jun. 2019, pp. 414–424. doi: 10.1145/3330345.3330382.

[57] A. Heinecke et al., "Understanding the Performance of Small Convolution Operations for CNN on Intel Architecture," SC'17, p. 2, 2017.

[58] S. N. Ragate, "Optimization of Spatial Convolution in ConvNets on Intel KNL," 2017.

[59] R. Gelashvili, N. Shavit, and A. Zlateski, "L3 Fusion: Fast Transformed Convolutions on CPUs," ArXiv191202165 Cs, Dec. 2019, Accessed: Sep. 19, 2021. [Online]. Available: http://arxiv.org/abs/1912.02165

[60] R. Wu, F. Zhang, Z. Zheng, X. Du, and X. Shen, "Exploring deep reuse in winograd CNN inference," in Proceedings of the 26th ACM SIGPLAN Symposium on Principles and Practice of Parallel Programming, Virtual Event Republic of Korea, Feb. 2021, pp. 483–484. doi: 10.1145/3437801.3441588.

[61] P. Maji, A. Mundy, G. Dasika, J. Beu, M. Mattina, and R. Mullins, "Efficient Winograd or Cook-Toom Convolution Kernel Implementation on Widely Used Mobile CPUs," in 2019 2nd Workshop on Energy Efficient Machine Learning and Cognitive Computing for Embedded Applications (EMC2), Washington, DC, USA, Feb. 2019, pp. 1–5. doi: 10.1109/EMC249363.2019.00008.

[62] H. Lan et al., "FeatherCNN: Fast Inference Computation with TensorGEMM on ARM Architectures," IEEE Trans. Parallel Distrib. Syst., vol. 31, no. 3, pp. 580–594, Mar. 2020, doi: 10.1109/TPDS.2019.2939785.

## AUTHORS

**GanTong** (1995- ) is a postgraduate of National University of Defense Technology. His main research area is CNN acceleration on different computer architectures.

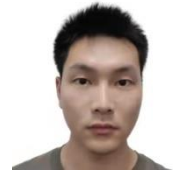

**Libo Huang** (1983- ) is a master supervisor of National University of Defense Technology. He is an associate researcher and mainly specializes in computer architecture.

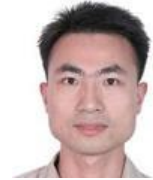